**Integration of nested cross-validation, automated hyperparameter optimization, high-performance computing to reduce and quantify the variance of test performance estimation of deep learning models**

Paul Calle[1], Averi Bates[1], Justin Reynolds[1], Yunlong Liu[1], Haoyang Cui[1], Sinaro Ly[1], Chen Wang[2], Qinghao Zhang[2], Alberto J. de Armendi[3], Shashank S. Shettar[3], Kar-Ming Fung[4,5], Qinggong Tang[2,5*], Chongle Pan[1,2*]

[1]School of Computer Science, University of Oklahoma, Norman, OK
[2]Stephenson School of Biomedical Engineering, University of Oklahoma, Norman, OK
[3]Department of Anesthesiology, University of Oklahoma Health Sciences Center, Oklahoma City, OK, 73104, USA.
[4]Department of Pathology, University of Oklahoma Health Sciences Center, Oklahoma City, OK 73104, USA.
[5]Stephenson Cancer Center, University of Oklahoma Health Sciences Center, Oklahoma City, OK 73104, USA.

*Correspondence: please contact cpan@ou.edu for questions on deep learning and contact qtang@ou.edu for questions on medical imaging.

## Abstract

*Background and Objectives*: The variability and biases in the real-world performance benchmarking of deep learning models for medical imaging compromise their trustworthiness for real-world deployment. The common approach of holding out a single fixed test set fails to quantify the variance in the estimation of test performance metrics. This study introduces NACHOS (Nested and Automated Cross-validation and Hyperparameter Optimization using Supercomputing) to reduce and quantify the variance of test performance metrics of deep learning models.

*Methods: NACHOS integrates Nested Cross-Validation (NCV) and Automated Hyperparameter Optimization (AHPO) within a parallelized high-performance computing (HPC) framework. NACHOS was demonstrated on a chest X-ray repository and an Optical Coherence*

*Tomography (OCT) dataset under multiple data partitioning schemes. Beyond performance estimation, DACHOS (Deployment with Automated Cross-validation and Hyperparameter Optimization using Supercomputing) is introduced to leverage AHPO and cross-validation to build the final model on the full dataset, improving expected deployment performance.*

*Results: The findings underscore the importance of NCV in quantifying and reducing estimation variance, AHPO in optimizing hyperparameters consistently across test folds, and HPC in ensuring computational feasibility.*

*Conclusions: By integrating these methodologies, NACHOS and DACHOS provide a scalable, reproducible, and trustworthy framework for DL model evaluation and deployment in medical imaging*. To maximize public availability, the full open-source codebase is provided at https://github.com/thepanlab/NACHOS

The variability and biases in the real-world performance benchmarking of deep learning models for medical imaging compromise their trustworthiness for real-world deployment. The common approach of holding out a single fixed test set fails to quantify the variance in the estimation of test performance metrics. This study introduces NACHOS (Nested and Automated Cross-validation and Hyperparameter Optimization using Supercomputing) to reduce and quantify the variance of test performance metrics of deep learning models. NACHOS integrates Nested Cross-Validation (NCV) and Automated Hyperparameter Optimization (AHPO) within a parallelized high-performance computing (HPC) framework. NACHOS was demonstrated on a chest X-ray repository and an Optical Coherence Tomography (OCT) dataset under multiple data partitioning schemes. Beyond performance estimation, DACHOS (Deployment with Automated Cross-validation and Hyperparameter Optimization using Supercomputing) is introduced to leverage AHPO and cross-validation to build the final model on the full dataset, improving expected deployment performance. The findings underscore the importance of NCV in quantifying and reducing estimation variance, AHPO in optimizing hyperparameters consistently across test folds, and HPC in ensuring computational feasibility. By integrating these methodologies, NACHOS and DACHOS provide a scalable, reproducible, and trustworthy framework for DL model evaluation and deployment in medical imaging. To maximize public availability, the full open-source codebase is provided at https://github.com/thepanlab/NACHOS

## 1. Introduction

Deep learning (DL) has achieved expert-level performance in numerous medical applications, including diabetic retinopathy detection from fundus photographs[1], skin cancer classification from dermoscopy images [2], and pneumonia detection in chest X-rays [3]. Despite these advances, the clinical deployment of DL systems remains limited due to concerns about generalizability, transparency, and regulatory readiness[4-6]. One central concern is the reliability of performance estimation: many DL studies report high accuracy using a single test split, but the variance and bias of these estimates are rarely assessed [7-12]. For example, Roberts et al. [13] found widespread methodological issues in COVID-19 diagnostic models, where overoptimistic performance claims stemmed from inadequate evaluation protocols.

In a typical deep learning study, a small fraction (i.e. 10%-20%) of the labeled data is held out in the test set for benchmarking the performance of the final model in unseen data, while the majority of the labeled data is used in the training and validation sets for model development. Relying on a single test split is akin to evaluating a school's academic performance based on one randomly chosen class—any performance estimate derived from it may be unrepresentative. In particular, the variance of the test performance metrics is typically unknown (because there is only one test set) and large (because only a small fraction of the labeled data is allocated to the test set). To make deep learning models trustworthy for medical decision-making, it is essential to estimate their performance metrics with low variance using more test data and measure the variance of the obtained estimates [14, 15].

The key objectives of this study are:

1. To demonstrate how nested cross-validation (NCV) provides reduced-variance and uncertainty-quantified estimates of deep learning model performance.
2. To integrate automated hyperparameter optimization (AHPO) within NCV using high-performance computing (HPC).
3. To introduce the NACHOS pipeline, which unifies NCV, AHPO, and HPC into a scalable framework for benchmarking.

4. To compare different data partitioning strategies in NCV and show their effects on test performance metrics.

5. To present DACHOS, a deployment-oriented algorithm that produces a final model optimized and trained on the full dataset

Nested cross-validation (NCV) is an effective procedure to meet this requirement. Briefly, the entire dataset is partitioned into *k* folds that are rotated through a cross-testing loop. A model development procedure with a (*k*-1)-fold cross-validation loop is nested within the cross-testing loop. The output of NCV is *k* estimates of the test performance metrics of *k* models. The average and variance of these *k* estimates reflect the expected performance and variability of this model development procedure across the entire dataset.

NCV has been used in a few medical machine learning studies. Nawabi et al. [16] employed NCV to benchmark the performance of a random forest classifier for prediction of survival for acute intracerebral hemorrhage using extracted radiomic features obtained from non-enhanced computed tomography images. Their random forest classifier achieved an average test accuracy of 72% with a 95% confidence interval between 70% and 74%. We utilized NCV to benchmark the performance of convolutional neural networks (CNNs) for analysis of Optical Coherence Tomography (OCT) images in multiple endoscopic applications [17-20]. For example, the average test classification accuracy of CNN was found to be 82.6% with 3.0% standard error for detecting three different renal tissues from their OCT images. However, NCV is still under-utilized in the medical field. Roberts et al. [13] conducted an analysis of COVID-19 research papers published between January and October 2020 and found a notable lack of NCV utilization, highlighting a gap in methodological rigor in the field.

A challenge of using NCV in a study is the need to implement automated hyperparameter optimization (AHPO) between the cross-testing loop and the cross-validation loop. Most medical deep learning studies perform manual hyperparameter optimization (MHPO). Practitioners can manually evaluate various model architectures, learning rates, regularization methods, and other hyperparameters based on cross-validation performance and select the configurations with the best validation performance to build the final model. However, it is impractical to perform MHPO independently and consistently in every test fold of NCV. Instead, AHPO needs to be performed during each testing iteration of the *k*-fold cross-testing loop in NCV to automatically identify the model configuration with the best cross-validation performance. AHPO within NCV provides reproducible model optimization and prevents inadvertent information leakage from the test set to the validation set during MHPO.

A second challenge of using NCV with AHPO is the need for significantly more computing than cross-validation with MHPO. Fortunately, the computation in NCV and AHPO can be readily partitioned by data folds for the cross-testing loop or the cross-validation loop and by model configurations for the AHPO loop. The folds can be distributed across many GPUs to compute in parallel. Thus, high-performance computing (HPC) can be used to complete NCV and AHPO within a reasonable amount of wall-clock time.

While many deep learning pipelines such as NiftyNet [21], TorchIO [22], DeepNeuro [23], and GaNDLF [24] support model training and evaluation for medical imaging, they lack integrated support for NCV, AHPO, and HPC—limiting their utility for generating robust, reproducible benchmarks needed in clinical AI. To address these limitations, we developed NACHOS (**N**ested and **A**utomated **C**ross-validation and **H**yperparameter **O**ptimization using **S**upercomputing) to integrate NCV and AHPO into a parallelized computational workflow on HPC. A repository of chest X-ray datasets from the TorchXRayVision library [25], along with an kidney OCT dataset, derived from Wang et al. [19], were used to demonstrate NACHOS. We compared different strategies for partitioning the two datasets into *k* folds in NCV. The results showed the significance of partitioning in benchmarking the test performance of deep learning models [26, 27].

The outcome of the NACHOS algorithm is a reduced-variance and uncertainty-quantified estimation of test performance of the models generated by this computational procedure. To build the final model for production use, we developed an algorithm named **D**eployment with **A**utomated **C**ross-validation and **H**yperparameter **O**ptimization using **S**upercomputing (DACHOS). DACHOS identifies the overall best model configuration using all the data for the AHPO and cross-validation and then uses this configuration to train a model using all the data. Because the final model for deployment was hyperparameter-optimized and trained using more data than the *k* models generated in the *k*-fold NCV, its test performance, although unknown, is expected to be better than the average test performance of the *k* NCV models. We also demonstrated DACHOS using the chest X-ray repository and kidney OCT dataset. While NACHOS is designed for rigorous performance evaluation through NCV and AHPO, DACHOS focuses on training a final model for deployment using all available.

## 2. Methodology

### 2.1. Overview of the NACHOS and DACHOS Frameworks

The proposed methodology introduces two complementary frameworks: NACHOS and DACHOS. Both are designed to automate and standardize performance benchmarking and model generation in medical imaging using deep learning. NACHOS (Nested and Automated Cross-validation and Hyperparameter Optimization using Supercomputing) focuses on performance evaluation through nested cross-validation combined with automated hyperparameter optimization (AHPO). It rotates all data folds through the test set to reduce variance and provides uncertainty-aware model performance estimates. DACHOS (Deployment with Automated Cross-validation and Hyperparameter Optimization using Supercomputing), in turn, uses the same automated pipeline to produce a final production model, trained on the full dataset with the best hyperparameters identified during cross-validation.

Both frameworks share a similar structure: they iterate over hyperparameter configurations and data folds in a controlled, distributed computing environment. The primary difference lies in their goals: NACHOS emphasizes performance estimation across folds, while DACHOS is designed to maximize model performance for real-world use by training on the entire dataset. The next sections detail the nested loop structures that implement these workflows.

### **2.2** NACHOS algorithm

The NACHOS algorithm comprises three nested loops: the cross-testing (CT) loop, the AHPO loop, and the cross-validation (CV) loop. First, the dataset *D* is divided into *k* folds: $F_0, F_1, \ldots, F_{k-1}$ . The CT loop iterates over $i \in I = \{0, 1, 2, \ldots, k-1\}$ , where the fold, $F_i$, is held out as the test set, and the remaining folds are used for training and validation. The AHPO loop then iterates over $j \in J = \{0, 1, 2, \ldots, n-1\}$, where $n$ is the number of hyperparameter configurations to be tried and each $h_j$ denotes the $j^{th}$ hyperparameter configuration. Within the CV loop, the index $m \in I - i$ is used to reserve the fold, $F_m$, for validation while the model is trained on the remaining $k-2$ folds. The model's performance on the validation fold $F_m$ is recorded as $v_m^j$. After completing cross-validation, the average validation performance—i.e. cross-validation performance— for each hyperparameter $h_j$, denoted as $\bar{v}^j$, is calculated. Once the AHPO loop is completed, the best-performing hyperparameter $h_{j^*}$ is selected based on the highest cross-validation performance. The model is then trained using $h_{j^*}$ on all folds except the test fold $F_i$ and evaluated on the test fold $F_i$ with the result recorded as $t_i$. . After completing all k iterations of the cross-testing loop, the test performance values $t_0$ , $t_1, \ldots, t_{k-1}$ are aggregated. The nested loop structure of the NACHOS algorithm is illustrated in Figure 1A, showing the cross-testing loop, AHPO loop, and cross-validation loop with their respective reserved folds. The average test performance $\bar{t}$ and its standard error (SE) are computed as: $\bar{t} = \frac{1}{k}\sum_{i=0}^{k-1} t_i$, $SE = \frac{\sigma_t}{\sqrt{k}}$ where $\sigma_t = \sqrt{\frac{1}{k-1}\sum_{i=0}^{k-1}(t_i - \bar{t})^2}$. For all experiments in this study, the performance metric used was classification accuracy, defined as: $Accuracy = \frac{Number\ of\ correct\ predictions}{Total\ number\ of\ predictions} = \frac{TP+TN}{TP+TN+FP+FN}$. Where TP, TN, FP, and FN denote true positives, true negatives, false positives, and false negatives. Using this metric, three types of accuracy were distinguished:

- Validation accuracy

  The accuracy obtained on a single validation fold $F_m$ during the cross-validation loop. It measures how well a model trained on k−2 folds generalizes to the reserved validation fold.

- Cross-validation accuracy
  For each hyperparameter configuration $h_j$, the cross-validation accuracy is the average of its validation accuracies across all validation folds, $v_m^j$. This value is used to identify the best-performing configuration, $h_{j^*}$.
- Test accuracy
  After selecting $h_{j^*}$, the model is retrained on all folds except the held-out test fold $F_i$ and evaluated on $F_i$. The resulting accuracy $t_i$ represents an unbiased estimate of generalization for that CT iteration.

These metrics provide a reduced-variance and uncertainty-quantified estimate of model performance using nested cross-validation.

In the current implementation of NACHOS, the AHPO loop used a random search algorithm [28] to sample n=9 hyperparameter configurations from a predefined search space. Each configuration was constructed by randomly selecting one value from each of the following sets:

- Batch size: {16, 32, 64, 128},
- Learning rate: {0.01, 0.001, 0.0001},
- Learning rate decay: {0.01, 0.001, 0.0001},
- Momentum: {0.5, 0.9, 0.99},
- Nesterov acceleration: {Yes, No}.
- Architecture: {ResNet50 [29], InceptionV3 [30], Xception [31]}.

A total of 9 configurations were randomly sampled (Table 1) and evaluated using cross-validation performance metrics. The number n=9 was selected as a trade-off between computational budget and performance coverage, allowing efficient exploration of the search space on the available high-performance computing (HPC) resources. For the experiments of this paper, the performance metric used was accuracy:

**2.3.** DACHOS algorithm

The DACHOS algorithm generates a production model, $M$, for deployment using AHPO and cross-validation. The dataset *D* is split into *k* folds for cross-validation. The AHPO loop iterates through $n$ hyperparameter configurations $h_j$, $j \in J = \{0, 1, 2, \ldots, n-1\}$, which should be the same as those used by NACHOS. The cross-validation loop iterates through $m \in I = \{0, 1, 2, \ldots, k-1\}$ to select the fold $F_m$ for validation and train the model on the remaining folds. The validation performance is recorded as $v_m^j$. After the $k$-fold cross-validation is completed for the hyperparameter configuration $h_j$, its average validation performance—i.e. cross-validation performance—is calculated as $\bar{v}^j$. Once cross-validation for all hyperparameter configurations is completed, the best-performing hyperparameter $h_{j^*}$ is selected based on its cross-validation performance. Finally, the production model, $M$, is trained using $h_{j^*}$ with the entire dataset *D*. The workflow of the DACHOS algorithm is summarized in Figure 1B, where the AHPO loop and cross-validation loop are used to select the optimal hyperparameter configuration prior to training the final production model. The DACHOS algorithm maximizes the performance of the production model, $M$, for deployment by using the entire dataset for AHPO and then using the entire dataset for model training. As in NACHOS, the performance metric used throughout the DACHOS pipeline was classification accuracy, computed using the same formula. The validation performances $v_m^j$ and their average $\bar{v}^j$ for each hyperparameter configuration were based on this metric. The final model $M$, trained on the entire dataset with the selected optimal hyperparameter $h_{j^*}$, inherits the configuration that achieved the highest average cross-validation accuracy (i.e. $\bar{v}^{j^*} = max_j \bar{v}^j$).

**2.4 Parallelization strategy**

The NACHOS and DACHOS algorithms were parallelized using a Python implementation of the Message Passing Interface (MPI) standard provided in the mpi4py [32] library. Both algorithms employed MPI point-to-point communication to enable direct interaction between parallel processes. The NACHOS algorithm distributes a total of *k**(*k*-1)**n* training tasks over *g* GPUs, where *k* is the number of folds for NCV, *n* is the number of hyperparameter configurations for AHPO, and *g* is the number of GPUs. The DACHOS algorithm parallelizes *k*n* training tasks over *g* GPUs.

When launched, the two algorithms create a manager process along with *g* worker processes, with each worker process assigned to a separate GPU. The manager process is responsible for assigning the tasks and sending their hyperparameter configurations, test folds, and validation folds to the worker processes for computing on their assigned GPUs. When a worker process completes a training task, it requests a new task from the manager process until all the tasks are completed. The dynamic scheduling provides effective load balancing and ensures linear scalability.

### 2.5 Fault tolerance Mechanism

To manage unexpected failures of training tasks in a job, NACHOS and DACHOS implement fault tolerance using a checkpointing system that includes two types of checkpoints: a metadata checkpoint and a model checkpoint. During training, the system continuously records the hyperparameter configuration $h_j$, the test fold $F_i$, the validation fold $F_m$, and the epoch number in the metadata checkpoint. After each epoch, a model checkpoint is saved while the previous one is deleted to conserve space. In the event of a failure, the job is relaunched and the manager process redistributes all training tasks. Each task corresponds to a unique combination of ($F_i$, $h_j$, $F_m$). Upon receiving a task, a worker process consults the metadata checkpoint to determine whether it has already been completed.

- If completed: the worker skips the task and requests a new one from the manager.
- If incomplete: the worker loads the latest model checkpoint (if available) and resumes training from the last saved epoch. Once finished, the worker reports completion and requests another task.

This process continues until the manager confirms that all tasks are finished. By coordinating checkpoints with task redistribution, NACHOS and DACHOS ensure that interrupted jobs resume without redundant computation or manual intervention. A schematic of this recovery mechanism is shown in Figure 2

### 2.6 Platform and dependencies

NACHOS and DACHOS were implemented using Python 3.8, TensorFlow 2.6.2, and scientific computing libraries. They support both distributed GPU workstations and HPC environments. Details regarding software dependencies, cluster configuration, and hardware specifications are provided in the Supplementary Section S1 to support reproducibility. The library can also generate class activation maps for instance-wide prediction interpretation or feature importance [33, 34] using GradCAM [35].

### 2.7 Medical imaging datasets and partitioning schemes

A comprehensive chest X-ray repository was assembled using four publicly available datasets included in the TorchXRayVision library [25]: the NIH ChestX-ray14 dataset (https://www.kaggle.com/datasets/nih-chest-xrays/data) [36], the CheXpert v1.0 dataset (https://stanfordmlgroup.github.io/competitions/chexpert/) [37], the MIMIC-CXR dataset (https://doi.org/10.6084/m9.figshare.10303823) [38], and the PadChest dataset (https://bimcv.cipf.es/bimcv-projects/padchest/) [39]. These datasets collectively provide a large and diverse collection of frontal-view chest radiographs labeled for various thoracic pathologies, making them suitable for benchmarking deep learning models in medical image analysis.

The NACHOS and DACHOS algorithms were evaluated on a binary classification task, in which deep learning models were trained to classify Posterior-Anterior (PA) chest X-ray images as either cardiomegaly or no finding. In the MIMIC-CXR and PadChest datasets, some lateral images were mistakenly labeled as PA. These incorrectly labeled images were identified through manual inspection and removed from our chest X-ray repository. All images were resized to a resolution of 224x224 pixels through interpolation. To create balanced data for benchmarking, we used a reproducible random selection (fixed seed) to choose 620 cardiomegaly images and 620 no-finding images from each of the four datasets. These images are combined to build the chest X-ray repository with a total of 4,960 images (4 datasets X 2 classes X 620 images per class per dataset). Consistent with the chest X-ray preprocessing approach, no contrast normalization or data augmentation was applied to the OCT dataset. The chest X-ray repository was partitioned into four folds using three different partitioning levels. In the image-level partitioning, images were randomly distributed across four folds. In the patient-level partitioning, all images from the same patient were assigned to the same fold. Finally, in the dataset-level partitioning, each dataset was exclusively allocated to a separate fold

An OCT dataset was derived from our previous study [18] for a renal tissue classification task. The OCT images were originally captured as 3D volumes, each containing multiple 2D cross-sectional B-scan images. These 2D cross-sectional images with a resolution of 185x210 pixels were used as the input data in this study. Similarly to chest X-ray repository, no contrast normalization nor data augmentation was performed. The OCT dataset contains 600 images of the cortex tissue, 600 images of the medulla tissue, and 600 images of the pelvis tissue from each kidney. A total of 10 kidneys were included, yielding 18,000 images (10 kidneys X 3 tissue types X 600 images per tissue type per kidney). The OCT dataset was partitioned into 10 folds at three levels: image, volume, and kidney. In the image-level partitioning, all 18,000 images were randomly split into 10 folds. In the volume-level partitioning, images from the same volume were assigned to the same fold. In kidney-level partitioning, the 10 kidneys were divided into 10 folds, with each fold containing all images from a respective kidney. The datasets used can be found in https://doi.org/10.5281/zenodo.14847200.

## 3. Results

### 3.1 Reduced-variance estimation of the test performance with uncertainty quantification using the NACHOS algorithm.

Table 3 presents the benchmarking results generated by the NACHOS algorithm on the chest X-ray repository for the cardiomegaly detection task. The data was partitioned into four folds corresponding to the four datasets. NACHOS included three nested loops: the CT loop, the AHPO loop, and the CV loop. For each test fold, the AHPO loop evaluated the cross-validation performance of nine hyperparameter configurations shown in Table 1. When $F_0$ was reserved as the test fold, configuration $h_2$ had the highest cross-validation accuracy of $\bar{v}^2 = 0.72$, which was the average of the validation accuracies.. After finding $h_2$ as the configuration with the best cross-validation accuracy, a model was trained on $F_1$, $F_2$, and $F_3$ using $h_2$ and then tested on $F_0$ to generate a test accuracy of $t_0$ =0.79. This process was repeated for the remaining test folds−$F_1$, $F_2$, and $F_3$. The standard deviation of the test accuracies was 0.06. The variability of the test performance stemmed from the different allocations of data folds for training, validation and testing, as well as the stochastic nature of stochastic gradient descend in model training and random search in AHPO.

The average test accuracy across all four folds was 0.75 with a standard error of 0.03, corresponding to a 95% t-based confidence interval of [0.67,0.84]This average test accuracy, derived from four model instances, reflected the expected performance of model instances that can be generated by our development procedure. The standard deviation of this average test accuracy (0.03) was half ($\sqrt{4}$) of the standard deviation of the individual test accuracies (0.06) from different test folds, because the average test accuracy represented the test performance from all four folds. From a practical application perspective, this variability implies that when the model is applied to new but similar clinical data, its accuracy would typically be expected to fall within about ±3% of the reported average.

The test accuracies were lower than the cross-validation accuracies in test folds $F_1$ and $F_2$ and higher in test folds $F_0$ and $F_3$. Although the AHPO was expected to make the cross-validation accuracies higher than the test accuracies, the results were inconsistent probably because of the data variability.

The average confusion matrix across the four test folds is shown in Table 5. On average, the models correctly classified 426 ± 82 of the 612 “No-Finding” cases and 505 ± 66 of the 612 “Cardiomegaly” cases. Misclassifications occurred in 194 ± 82 “No-Finding” images predicted as “Cardiomegaly” and 115 ± 66 “Cardiomegaly” images predicted as “No-Finding.” These results suggest that false positives for cardiomegaly were more frequent than false negatives, which could have implications for clinical workflows—potentially leading to more follow-up examinations but reducing the likelihood of missed diagnoses. The standard deviations reflect moderate variability across folds, indicating that performance on new datasets from similar imaging sources would likely remain within the reported ranges.

The results from the NACHOS algorithm on the kidney OCT dataset are summarized in Table 4. The OCT images were partitioned into 10 folds with each fold containing all the images from one kidney. When $F_0$ was reserved as the test set, hyperparameter configuration $h_5$ yielded the best cross-validation accuracy of 0.93, which was the average of 9 validation accuracies from the 9-fold cross-validation from $F_1$ to $F_9$. Then, a model trained on the nine folds from $F_1$ to $F_9$ using configuration $h_5$ achieved a test accuracy of $t_0$ =0.73 on the reserved test fold $F_0$. The cross-testing loop in NACHOS repeated this process for the remaining nine folds. The test accuracies had a wide range from 0.71 to 0.96 with a standard deviation of 0.09, which reflected the significant data variability from kidney to kidney in this dataset.

The average test accuracy from the 10 test folds was 0.84 with a standard error of 0.03, corresponding to a 95% t-based confidence interval of [0.77,0.90]. The cross-testing procedure reduced the standard deviation of the average test accuracy estimation by approximately three folds ($\sqrt{10}$) from 0.09 to 0.03 by averaging across 10 individual test accuracies from separate kidney samples. The standard deviation of the individual folds' test accuracy was higher in the kidney OCT data (0.09) than the chest X-ray data (0.06), but the standard error of the average test accuracy was 0.03 in both datasets because of the higher number of test folds used in the kidney OCT data than the chest X-ray data. From a practical standpoint, this indicates that when the model is applied to new kidney samples in a clinical context, performance can be expected to vary by roughly ±3% around the mean accuracy, reflecting moderate but manageable variability in generalization across patients.

---

The average confusion matrix across the ten test folds is shown in Table 6. The models achieved high accuracy but with notable differences in per-class performance and variability across folds. Cortex was correctly identified in 454 ± 136 of 600 cases, with the most frequent confusion being misclassification as Pelvis (106 ± 147), indicating that these two tissues share imaging characteristics that can occasionally lead to errors. Medulla achieved a high correct classification rate (512 ± 137 of 600), with very few false predictions as Pelvis (8 ± 19) and some misclassification as Cortex (80 ± 136), suggesting strong discriminative features for this class. Pelvis was correctly classified in 543 ± 76 of 600 cases, with most of its misclassifications occurring as Cortex (55 ± 72). The relatively large standard deviations reflect variability in model performance across different kidneys, likely due to inter-kidney anatomical differences and imaging conditions. Clinically, these results suggest that while Medulla and Pelvis classifications are generally robust, distinguishing Cortex from Pelvis in new kidney data remains more challenging, and further refinement or additional training data may help improve consistency.

Figure 3 shows Grad-CAM results for cross-testing, highlighting the regions most influential in the model's decision, for test fold 7 on Pelvis images. Figure 3A illustrates a correct classification, while Figure 3B shows an incorrect one. The primary difference in the red-highlighted region is that Figure 3A contains a greater concentration of bright speckle clusters, whereas Figure 3B appears smoother and more uniform, which likely contributed to the misclassification.

### 3.2 Development of production model for deployment using the DACHOS algorithm.

The DACHOS algorithm was used to produce a final model for deployment. The results of applying DACHOS to the chest X-ray repository are shown in Table 4. When hyperparameter configuration $h_0$ was used, the validation accuracies were $v_0^0 = 0.71$ for fold $F_0$, $v_1^0 = 0.71$ for fold $F_1$, $v_2^0 = 0.69$ for fold $F_2$, and $v_3^0 = 0.70$ for fold $F_3$, which yielded a cross-validation accuracy of $\bar{v}^0 = 0.71$ for $h_0$. DACHOS and NACHOS produced different validation accuracies for the same configuration, because DACHOS used 3 data folds for training while NACHOS used only 2. Four-fold cross-validation was performed for all remaining hyperparameter configurations; configuration $h_5$ achieved the highest cross-validation accuracy of $\bar{v}^5 = 0.75$. This configuration was then used to train a final model on all four data folds.

For the same configuration, 78% of the cross-validations accuracies for NACHOS were lower than the cross-validation accuracy for DACHOS. For example, with hyperparameter configuration $h_5$, NACHOS achieved cross-validation accuracies of 0.72, 0.73, 0.71, and 0.69 for test folds $F_0$ to $F_3$ using 2 data folds for training. In contrast, DACHOS achieved a cross-validation accuracy of 0.75 with $h_5$ using 3 data folds for training. An additional data fold available for training in DACHOS may have contributed to its increased cross-validation accuracy. .

The DACHOS algorithm was also applied to the kidney OCT dataset (Table 5). Here, hyperparameter configuration $h_0$ achieved validation accuracies of 0.63, 0.88, 0.85, 0.76, 0.91, 0.97, 0.92, 0.87, 0.94, 0.88 for folds $F_0$ to $F_9$, which was averaged to a cross-validation accuracy of $\bar{v}^0 = 0.86$ . This process was repeated for the remaining hyperparameter configurations. Hyperparameter configuration $h_2$ had the highest average accuracy $\bar{v}^2 = 0.90$ and $h_3$ had the lowest $\bar{v}^3 = 0.84$. The production model was trained using $h_2$ on the entire dataset.

### 3.3 Evaluation and interpretation of model performance across different partitioning levels.

NACHOS and DACHOS required data to be partitioned into multiple folds for NCV and CV. An appropriate partitioning design was essential for evaluating a model's ability to generalize to unseen data from a new measurement, a new patient or a new location. The input data in the chest X-ray repository was organized into three partitioning levels, including the image level, the patient level, and the dataset level, for testing to evaluate different partitioning designs (Figure 4A). The chest X-ray repository comprised a total of 4,960 images from 4,678 patients across four datasets. There were 1,187 patients in the CheXpert dataset, 1,165 patients in the MIMIC-CXR dataset, 1,105 patients in the ChestX-ray8 dataset, and 1,221 patients in the PadChest dataset. The dataset-level partitioning was used in the previous Results sections to evaluate the model performance on unseen data from a different location. Here, we compared the test performance benchmarked using NACHOS across image-level, patient-level, and dataset-level partitioning (Figure 4B). Because each patient had approximately 1.06 images on average in the chest X-ray repository, the image-level and the patient-level partitioning yielded similar average test accuracies of 0.811 and 0.809, respectively, which reflected the test performance of the models on unseen data from new images or new patients within the 4 datasets. These were much higher than the average test accuracy of 0.750 from the dataset-level partitioning (Figure 4B). The variability in test accuracies across data folds also increased from 0.008 for the image-level partitioning and 0.009 for the patient-level partitioning to 0.061 for the dataset-level stratification. A one-way ANOVA found no statistically significant difference in accuracy distributions among the three strategies ($F = 3.70$, $p = 0.067$), with variance attributed to partitioning (treatment) estimated at 0.00088 and residual variance at 0.001297. However, pairwise effect size analysis indicated practically meaningful differences: the contrast between image-level and patient-level was small (Cohen's $d = 0.24$), whereas comparisons involving dataset-level were large (image vs. dataset: $d = 1.40$; patient vs. dataset: $d = 1.35$). These results suggest that while statistical significance was not reached at the conventional 0.05 threshold, dataset-level partitioning had a substantial practical impact, markedly reducing test accuracy compared with the other strategies. From a clinical reliability perspective, the low fold-to-fold variability at the image and patient levels indicates that model performance is relatively stable when tested on data similar to the training set, which increases confidence in consistent results within the same clinical setting. In contrast, the larger variability at the dataset level reflects reduced predictability when applying the model to data from new sites, underscoring the need for external validation before deployment in different clinical environments.

The test performances were also benchmarked using NACHOS on the kidney OCT dataset at three partitioning levels: image, volume, and kidney (Figure 5A). The kidney-level partitioning was used in the previous Results sections to evaluate the model performance on unseen data from a new kidney. Here, we compared the test performance benchmarked by NACHOS using the image-level, volume-level, and kidney-level partitioning (Figure 5B). The image-level partitioning resulted in a perfect test accuracy of 1.00 across all data folds owing to nearly complete data redundancy among contiguous images from the same volume. The volume-level partitioning generated an average test accuracy of 0.97 and a standard deviation of 0.01 across different data folds, suggesting still substantial data redundancy among volumes from the same kidney. In contrast, the kidney-level partitioning yielded an average test accuracy of 0.84 and a standard deviation of 0.09. The reduced accuracy and increased variability reflected the real-world variability of the OCT data from different kidneys and the need for the models to generalize well across kidneys. A one-way ANOVA was conducted to compare mean accuracies across the three splits. The analysis indicated a significant effect of split on accuracy, $F(2,27)=27.08$, $p=3.52*10^{-7}$. The between-split variance component was estimated at 0.007043, and the within-split (error) variance component at 0.002700. Post-hoc comparisons using Tukey's HSD revealed that: Split 1 vs Split 3: Split 1 (M = 0.811) had significantly higher accuracy than Split 3 (M = 0.750), mean difference = 0.162, 95% CI [0.104, 0.219], $p < 0.001$. Split 2 vs Split 3: Split 2 (M = 0.809) had significantly higher accuracy than Split 3 (M = 0.750), mean difference = 0.130, 95% CI [0.072, 0.187], $p < 0.001$. Split 1 vs Split 2: No significant difference was found, mean difference = 0.032, 95% CI [–0.026, 0.090], p = 0.367. Effect size analysis showed that these differences were not only statistically significant but also large in magnitude: image-level vs. volume-level (d=3.94), image-level vs. kidney-level (d=2.56), and volume-level vs. kidney-level (d=2.04), underscoring the substantial practical impact of partitioning choice on OCT model performance. From a clinical reliability perspective, the very low fold-to-fold variability at the image and volume levels suggests that the model would produce consistent results when applied to data that is highly similar to its training set. In contrast, the markedly higher variability at the kidney level indicates that performance can fluctuate substantially when encountering entirely new patient samples, highlighting the importance of validating the model across diverse clinical scenarios before broader adoption.

### 3.4 Parallelization of NACHOS and DACHOS over multiple GPUs

We compared the execution time of the NACHOS algorithm on various GPU systems using the kidney OCT dataset with kidney-level partitioning using a single hyperparameter configuration. These systems included a Beowulf cluster of GPU workstations with Nvidia GeForce RTX 4090 or Nvidia RTX A6000 GPUs on a local Ethernet network, as well as the OSCER supercomputer with Nvidia A100 GPUs. The execution time was 21.9 hours on a single RTX A6000, 13.8 hours on a single RTX 4090, and 11.7 hours on a single A100 (Figure 6A). RTX A6000 has 336 tensor cores and memory bandwidth of 112.5 GB/s, RTX 4090 has 512 tensor cores and memory bandwidth ~1000 GB/s, and A100 has 432 tensor cores and memory bandwidth of ~2000 GB/s. RTX 4090 performed better than RTX probably due to higher number of tensor cores. A100 performed better than RTX 4090 probably due to higher memory bandwidth as well as optimizations for deep learning applications. The peak memory usages on these GPUs were approximately all 18.5 GB. This demonstrated the portability of NACHOS and DACHOS across a variety of computing systems.

NACHOS and DACHOS can distribute the training across multiple GPUs with fault tolerance to reduce the wall-clock time. Figure 6B presents the speedup ratios by the number of GPUs using the execution time of a single RTX A6000 GPU (21.9 hours) as the baseline. NACHOS reached linear scalability on RTX A6000s. The execution time was reduced to 11.1 hours on 2 RTX A6000s with a 2.0X speedup, further to 6.9 hours on 3 RTX A6000s with a 3.2X speedup, and finally to 5.4 hours on 4 RTX 6000s with a 4.1X speedup. NACHOS achieved super-linear speedups on RTX 4090s and A100s. The execution time was 13.8 hours on 1 RTX 4090, 7.0 hours on 2 RTX 4090s (3.1X speedup), and 4.7 hours on 3 RTX 4090s (4.6X speedup), and 3.6 hours on 4 RTX 4090s (6.1X speedup). Super-linear speedups were also achieved on A100s: 3.7X on 2 A100s, 7.4X on 4 A100s, and 14.1X on 8 A100s. The execution time was reduced to only 1.6 hours using 8 A100s.

## 4. Discussion

Accurate and robust benchmarking of the performance of machine learning models has been a challenge in the field of medical imaging [40]. A commonly used approach is to split a labeled dataset into a training set for learning model parameters, a validation set for optimizing model hyperparameters, and a test set for benchmarking the performance of the obtained model. Although cross-validation is often used to rotate data partitions between the training set and the validation set, most machine learning studies split out a single fixed test set for performance benchmarking.

NACHOS features NCV in an automated and user-friendly machine learning workflow for medical imaging applications. *k*-fold NCV offers two key advantages over a single test split for test performance benchmarking. First, NCV reduces the variance of the test performance estimation by rotating all data partitions through the test set.

Specifically, the variance of the average performance score from *k*-fold NCV should be *k* times lower than the variance of the point estimate of the performance score from a single test split. Second, and more importantly, the variance of the performance estimation is not quantified using a single test split, but is quantified during cross-testing over k test partitions. For example, if a user holds out only the last partition for the test set, they would estimate the classification accuracy to be 0.82 in the chest X-ray repository and 0.86 in the kidney OCT dataset with large (± 0.06 and ± 0.09, respectively) variabilities that are unknown to the user. If the user uses NVC, they would be able to better estimate the accuracy as 0.75 ± 0.03 in the chest X-ray repository and as 0.84 ± 0.03 in the kidney OCT dataset. These use cases demonstrated that NCV reduced and measured the variance of the performance benchmarking of deep learning models.

The partitioning level used by NCV is also important for accurate and robust performance benchmarking. The Checklist for Artificial Intelligence in Medical Imaging (CLAIM) [41] emphasizes transparent reporting of data partitioning and recommends partitioning at the patient level or higher. The classification accuracy in the chest X-ray repository decreased from 0.809 with the patient-level partitioning to 0.750 with the dataset-level partitioning (Figure 4B). This means that it is more difficult for models to generalize to a new dataset acquired by other institutions than to generalize to new patients from a previously seen dataset. Similarly, Zech et al. [42] found that pneumonia classification models often perform better on internal test datasets originated from the same institution as the training data than on external test datasets from institutions different that the ones used for training. The choice of the partitioning level for NCV performance benchmarking should match the intended use scenario of the models. For instance, the patient-level partitioning can be used to evaluate models designed for use within the same hospitals that produced the training data, as it mimics the scenario of encountering new patients within these hospitals. The institution-level partitioning should be used to evaluate models intended to be deployed to new hospitals.

NACHOS benchmarks the average test performance of models generated by a reproducible workflow using a specific dataset. AHPO is needed in NACHOS because it is impractical to perform laborious manual hyperparameter optimization consistently across all test folds in NCV. Random search [28] is a simple, yet highly effective, AHPO method. In the chest X-ray repository (Table 2), the best hyperparameter configuration achieved improvements ranging from 0.07 to 0.10 over the average cross-validation accuracy of all configurations. In the kidney OCT dataset (Table 3), AHPO delivered performance gains between 0.02 and 0.05 above the overall average.

NCV and AHPO in NACHOS incur a significant computational cost that needs to be distributed across multiple GPUs to shorten the wall-clock time of model development. The parallelization in NACHOS achieved linear and super-linear speedups on different kinds of GPUs (Figure 6A and 6B). For example, distributing training across four RTX A6000 GPUs reduced runtime from 21.9 to 5.4 hours—representing a 75% reduction in runtime to a single GPU. Even greater efficiency gains were observed on RTX 4090s and A100s, where super-linear scalability was achieved. On A100s, runtime decreased from 11.7 hours on a single GPU to just 1.6 hours on eight GPUs—an 86% reduction in runtime. These results underscore not only the portability of NACHOS across heterogeneous GPU architectures but also its strong scalability and efficiency for large-scale medical AI workflows. Super-linear speedup, observed in certain cases, suggests additional efficiency gains arising from improved cache utilization, reduced memory bottlenecks, or synergistic effects in GPU parallelism. After NACHOS measures the test performance of models from a model development workflow, DACHOS is used to generate a production model for deployment using this workflow. The actual test performance of this production model is unknown but should be higher than the test performance of the models benchmarked by NACHOS. This is because the AHPO in DACHOS can use an extra data fold than the AHPO in NACHOS and the final model training in DACHOS can use two additional data folds than the training in NACHOS.

Beyond performance benchmarking, NACHOS provides built-in support for interpretability through automated generation of class activation maps and receiver operating characteristic (ROC) curves, offering insight into both how a model makes predictions and where errors may occur. Such interpretability features are increasingly recognized as essential for explainable AI in healthcare, making them a core benefit of adopting NACHOS in addition to its variance-aware benchmarking capabilities.

In addition to variance-aware benchmarking and hyperparameter optimization, NACHOS incorporates engineering features that make it suitable for large-scale medical AI workflows. The framework demonstrated both linear and super-linear GPU scalability, ensuring that computationally intensive NCV and AHPO processes can be executed efficiently across modern HPC resources. Furthermore, NACHOS includes a fault-tolerant checkpointing system that safeguards progress during long training runs, enabling workflows to resume seamlessly after interruptions. Together, these capabilities underline the framework's robustness and practicality for large-scale, distributed model development in medical imaging.

### 4.1. Limitations

This study has several limitations that warrant consideration. First, we evaluated NACHOS and DACHOS only on chest X-ray and OCT datasets, and focused exclusively on image classification tasks. The framework needs to be further evaluated for other modalities such as CT or MRI, and for additional tasks such as segmentation and detection. While the principles of nested cross-validation, automated hyperparameter optimization, and distributed parallelization are broadly applicable, empirical validation across diverse imaging types and tasks will be necessary to confirm the generalizability of the framework.

Second, although nested cross-validation provides a strong foundation for variance-aware performance estimation, the study did not include evaluation on fully independent, external datasets. For the chest X-ray repository, our use of dataset-level partitioning partially addresses this by rotating test folds across data originating from different institutions, thereby simulating cross-site generalization. However, we acknowledge that this approach does not replace validation on entirely unseen, external datasets. Such validation across independent multi-institutional cohorts will be an important next step to more rigorously establish the robustness and real-world reproducibility of the framework.

Third, we did not conduct a quantitative comparison of NACHOS against existing machine learning pipelines such as NiftyNet or TorchIO. Such comparisons could provide more direct evidence of practical improvements in performance benchmarking, computational efficiency, and reproducibility. Our emphasis here was on introducing the unified integration of nested cross-validation, automated hyperparameter optimization, and high-performance computing within a single workflow, and demonstrating its feasibility across two representative datasets.

Fourth, while our results demonstrate scalability and portability across GPU platforms, the framework's dependence on multi-GPU setups and high-performance computing clusters introduces a barrier to adoption in lower-resource settings. This reflects a tradeoff between the reproducibility and robustness gained from nested cross-validation with automated hyperparameter optimization and the computational burden required to execute these workflows at scale. Broader adoption may require adaptations such as more efficient AHPO strategies, or cloud-based solutions to lower the entry barrier for institutions with limited infrastructure.

Fifth, while the framework is intended to support deployment in clinical workflows, this study does not present clinical validation or comparison against human expert performance. As such, the effectiveness, safety, and practical utility of models developed through NACHOS and DACHOS in real clinical environments remain unproven. Future work should include pilot clinical studies and direct benchmarking against human experts to more fully establish the framework's role in supporting trustworthy clinical AI adoption.

Sixth, the framework assumes that training and evaluation datasets contain accurate, noise-free labels. In practice, medical datasets often exhibit label noise and interobserver variability, especially in complex or ambiguous cases. Such label inconsistencies can influence both model performance and the estimated variance across folds While the reproducibility and variance quantification benefits of nested cross-validation still hold under moderate label noise, future work should explore strategies for robustness, such as incorporating consensus labeling, or integrating noise-aware training methods [43].

6. Conclusion

NACHOS integrates NCV, AHPO and HPC into an automated workflow. NCV reduces and measures variance in test performance estimation by rotating all data folds through the test set. This variance reduction applies to the estimation process rather than guaranteeing lower variability when the model is applied to new, external clinical environments. AHPO enhances model performance by searching for optimal hyperparameters and avoids the impracticality of manual tuning across multiple test folds in NCV. To mitigate substantial computational costs of NCV and AHPO, NACHOS can distribute computation across multiple GPUs with linear or super-linear speed-up. Although validated here on two imaging modalities (chest X-ray and OCT) and classification tasks, its design is general and can be extended to other medical imaging modalities and tasks such as segmentation. To our knowledge, this is the first framework to unify nested cross-validation, automated hyperparameter optimization, and high-performance computing for medical imaging. While each of these components has been applied individually in prior studies, their integration in NACHOS represents a novel and impactful advancement. This combination not only ensures variance-aware performance benchmarking and reproducible optimization but also makes such workflows computationally feasible at scale, thereby addressing a key gap in current medical AI development pipelines.

Following benchmarking, DACHOS can be used to generate final models that may achieve higher accuracy. However, these models should be viewed as technically optimized research outputs, not as clinically validated tools ready for deployment. While NCV reduces estimation variability, some performance variation is inherent to dataset characteristics and may still be observed when applying models to new clinical settings.

By providing transparent, reproducible, and scalable evaluation, NACHOS and DACHOS help address reproducibility concerns in medical AI and could inform best practices for model validation. Importantly, the entire codebase (v1.0.0) is openly available at https://github.com/thepanlab/NACHOS, enabling broad community adoption, extension to other domains, and independent verification of results—thereby maximizing transparency and accelerating progress beyond this study's scope.

Beyond methodological contributions, NACHOS and DACHOS address critical gaps in current development pipelines that directly impact regulatory trust and reproducibility in clinical settings. By standardizing performance benchmarking, quantifying estimation variance, and enabling transparent validation across data partitions, the framework provides a foundation for more reliable evaluation of medical AI systems. These qualities are essential not only for advancing research reproducibility but also for informing best practices and regulatory standards that govern the safe and responsible integration of AI into healthcare.

Looking ahead, several extensions of this work are planned. First, NACHOS will be expanded to support additional AHPO strategies, such as Hyperband [44] and Bayesian Optimization Hyperband [45], to further improve search efficiency. Second, the framework will be applied to a broader range of medical imaging modalities and tasks, including segmentation and multi-label classification. Finally, integration into pilot clinical workflows will be explored to evaluate practical usability, interoperability with existing hospital systems, and potential barriers to real-world adoption.

**Data availability**

The repository for NACHOS and DACHOS is available at https://github.com/thepanlab/NACHOS.

**Code availability**

The kidney OCT dataset and chest X-ray repository are available at https://doi.org/10.5281/zenodo.14847200.

**Acknowledgements**

Special thanks to Jessica Shaw for her assistance with the coding aspects of this project. This work was supported by grants from the University of Oklahoma Health Sciences Center (3P30CA225520), National Science Foundation (OIA-2132161, 2238648, 2331409), National Institute of Health (R01DK133717), Oklahoma Center for the Advancement of Science and Technology (HR23-071), the medical imaging COBRE (P20 GM135009), the Prevent Cancer Foundation, and the Midwest Biomedical Accelerator Consortium (MBArC), an NIH Research Evaluation and Commercialization Hub (REACH). Histology service provided by the Tissue Pathology Shared Resource was supported in part by the National Institute of General Medical Sciences COBRE Grant P20GM103639 and National Cancer Institute Grant P30CA225520 of the National Institutes of Health. Financial support was provided by the OU Libraries’ Open Access Fund.

**Author contributions:**

P.C. and C.P. designed the algorithms. P.C, implemented the algorithms. P.C, A.B. run the experiments. P.C., J.R., Y.L., H.C., S.S analyzed the results. C.P. and Q.T. served as the principal investigators of the funding sources. A.B. retrieved X-ray data. A.J. de A. and S.S.S. aided in OCT experiment design and provided clinical perspectives for the OCT data. K.M.F. aided in OCT experiment design and provided histology service to validate the tissue types. C.W and Q.Z. obtained the OCT data. P.C. and C.P. drafted the manuscript. All authors contributed to critical revisions and approved the final manuscript.

**Competing interests:**

The authors declare no competing interests.

**Figure 1: Schematic comparison of the NACHOS and DACHOS algorithms.** [A] NACHOS uses a nested structure with three loops: the cross-testing loop reserves one fold for testing (blue), the AHPO loop explores different hyperparameter configurations, and the cross-validation loop evaluates each configuration by rotating the validation fold (orange). The best configuration is retrained on all folds except the test fold and evaluated on the held-out test fold. [B] DACHOS employs AHPO and cross-validation without an outer cross-testing loop. Hyperparameter configurations are compared based on average cross-validation performance (orange), and the best configuration is used to retrain the final production model on the entire dataset.

**Figure 2: Fault-tolerant workflow for distributed training.** [A] The manager process creates a list of tasks and assigns them iteratively to available workers until all tasks are completed. [B] The worker process reports its availability, receives a task, and checks metadata to determine completion status. If the task is finished, it is skipped; otherwise, the worker loads the latest model checkpoint and resumes training. Afterward, the worker reports availability for reassignment. This coordination ensures efficient scheduling, prevents redundant computation, and enables recovery from failures.

**Figure 3: Grad-CAM explainability results for cross-testing on OCT dataset for fold k7.** [A] Correct classification [B] Incorrect Classification.

**Figure 4: Data partitioning schemes of the chest X-ray repository** [A] Data structure of the chest X-ray repository. The repository includes four datasets, each originating from a different set of institutions, capturing variations in imaging protocols and patient populations. [B] Boxplots. Mean ($\mu$) and standard deviation (SD) of the test accuracy from data partition on the image level, the patient level, and the dataset level. The four diamonds represent the test accuracies of four partitions of mixed images regardless of patients, four partitions of mixed patients regardless of their source datasets, and four

partitions corresponding to the four datasets. The dataset-level partitioning had significantly lower mean and higher variability in the test accuracies.

**Figure 5: Data partitioning schemes of the kidney OCT dataset.** [A] Data structure of the kidney OCT dataset. The dataset includes 10 kidneys, each generating 90 OCT volumes. 20 B-scan images are extracted from each volume. [B] Boxplots. Mean ($\mu$) and standard deviation (SD) of the test accuracy from data partition on the image level, the volume level, and the kidney level. The 10 diamonds represent the test accuracies of 10 partitions of mixed images regardless of volume, 10 partitions of mixed volume regardless of their source kidney, and 10 partitions corresponding to the 10 kidneys. The kidney-level partitioning had significantly lower mean and higher variability in the test accuracies.

**Figure 6: Scalability of NACHOS parallelization**. [A] Execution time as a function of the number of GPUs for three GPU types: RTX A6000 (Beowulf cluster), RTX 4090 (Beowulf cluster), and A100 (supercomputer). [B] Speedup relative to the number of GPUs. The speedup of all GPU types is calculated using the execution time of a single RTX A6000 GPU as the baseline (1X). NACHOS achieved linear speedup on the A6000 GPUs and super-linear speedup on the RTX 4090 GPUs and the A100 GPUs.

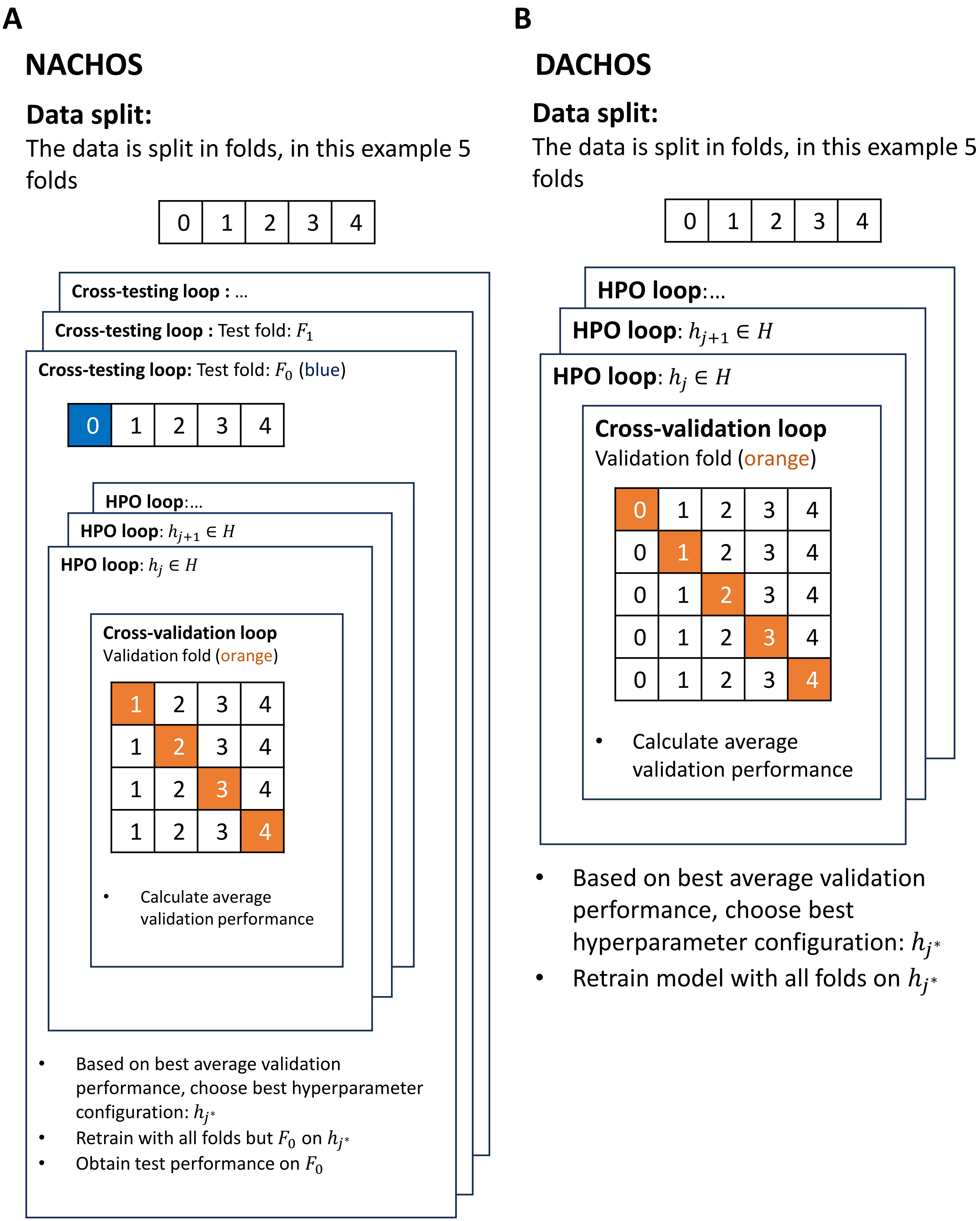
A
NACHOS
Data split:
The data is split in folds, in this example 5 folds
0 1 2 3 4
Cross-testing loop : ...
Cross-testing loop : Test fold: $F_1$
Cross-testing loop: Test fold: $F_0$ (blue)
0 1 2 3 4
HPO loop:...
HPO loop: $h_{j+1} \in H$
HPO loop: $h_j \in H$
Cross-validation loop
Validation fold (orange)
1 2 3 4
1 2 3 4
1 2 3 4
1 2 3 4
• Calculate average validation performance
• Based on best average validation performance, choose best hyperparameter configuration: $h_{j^*}$
• Retrain with all folds but $F_0$ on $h_{j^*}$
• Obtain test performance on $F_0$
B
DACHOS
Data split:
The data is split in folds, in this example 5 folds
0 1 2 3 4
HPO loop:...
HPO loop: $h_{j+1} \in H$
HPO loop: $h_j \in H$
Cross-validation loop
Validation fold (orange)
0 1 2 3 4
0 1 2 3 4
0 1 2 3 4
0 1 2 3 4
0 1 2 3 4
• Calculate average validation performance
• Based on best average validation performance, choose best hyperparameter configuration: $h_{j^*}$
• Retrain model with all folds on $h_{j^*}$

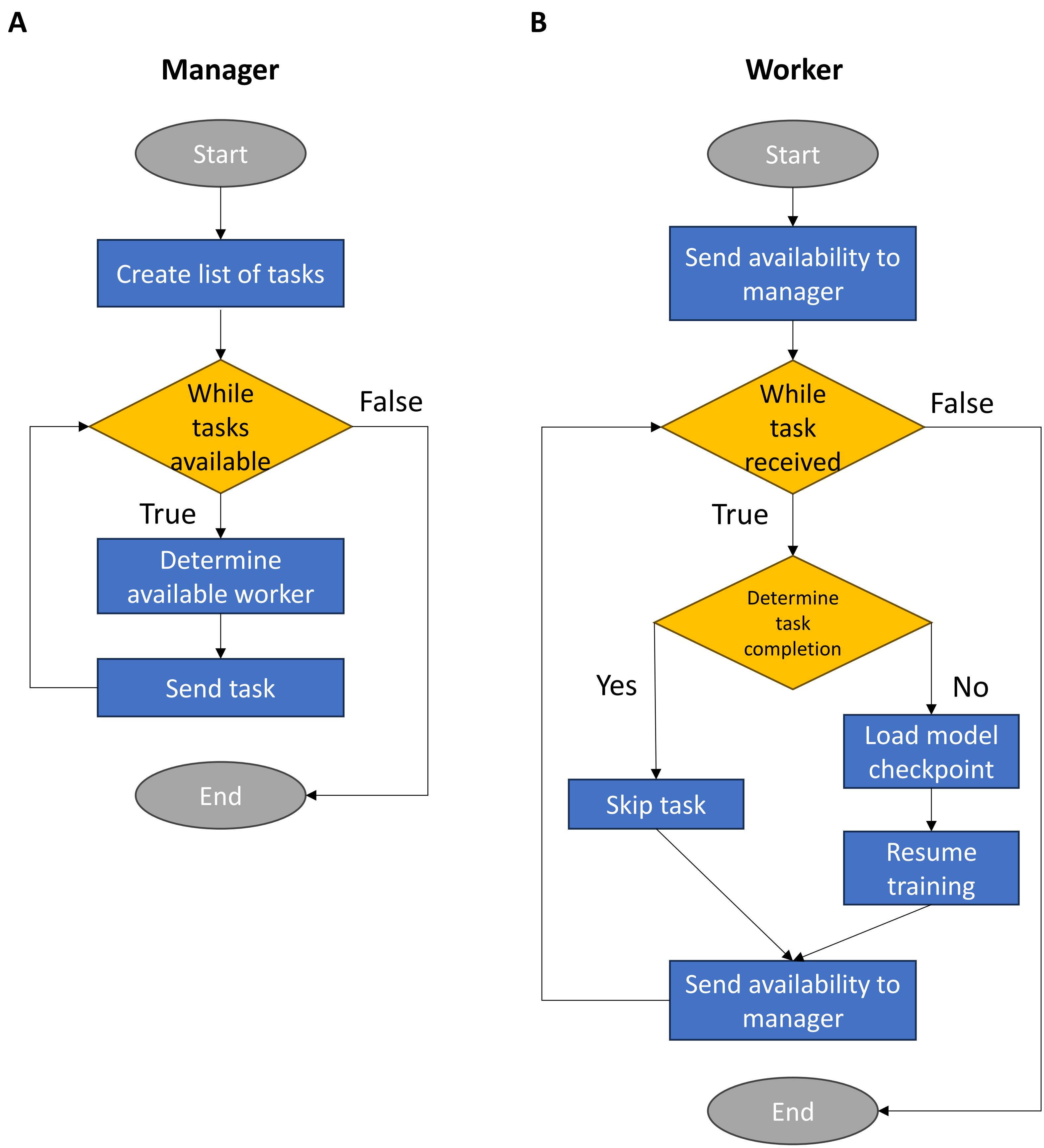
A
Manager
Start
Create list of tasks
While tasks available
False
True
Determine available worker
Send task
End
B
Worker
Start
Send availability to manager
While task received
False
True
Determine task completion
Yes
No
Load model checkpoint
Skip task
Resume training
Send availability to manager
End

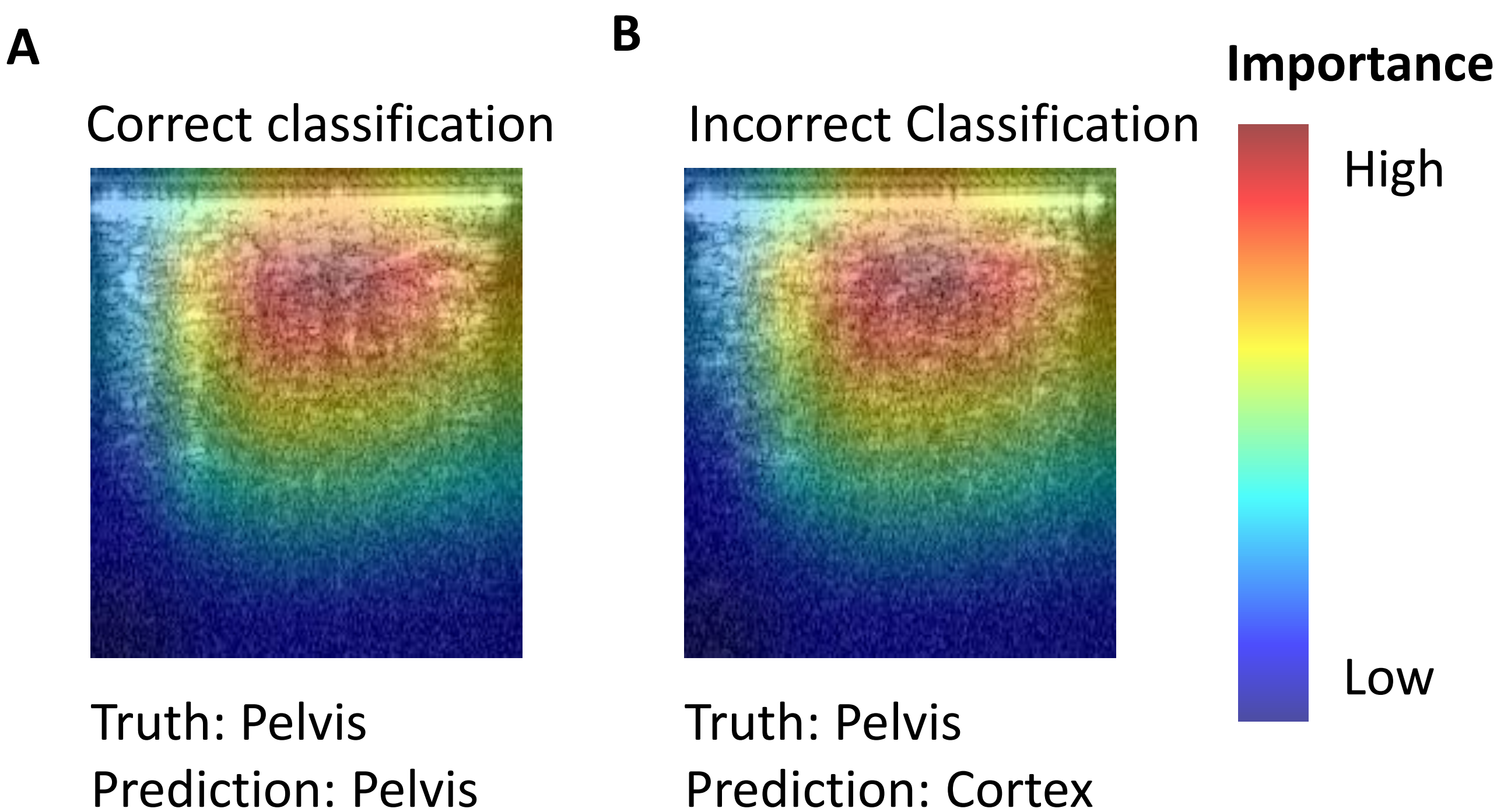
A
Correct classification
Truth: Pelvis
Prediction: Pelvis
B
Incorrect Classification
Truth: Pelvis
Prediction: Cortex
Importance
High
Low

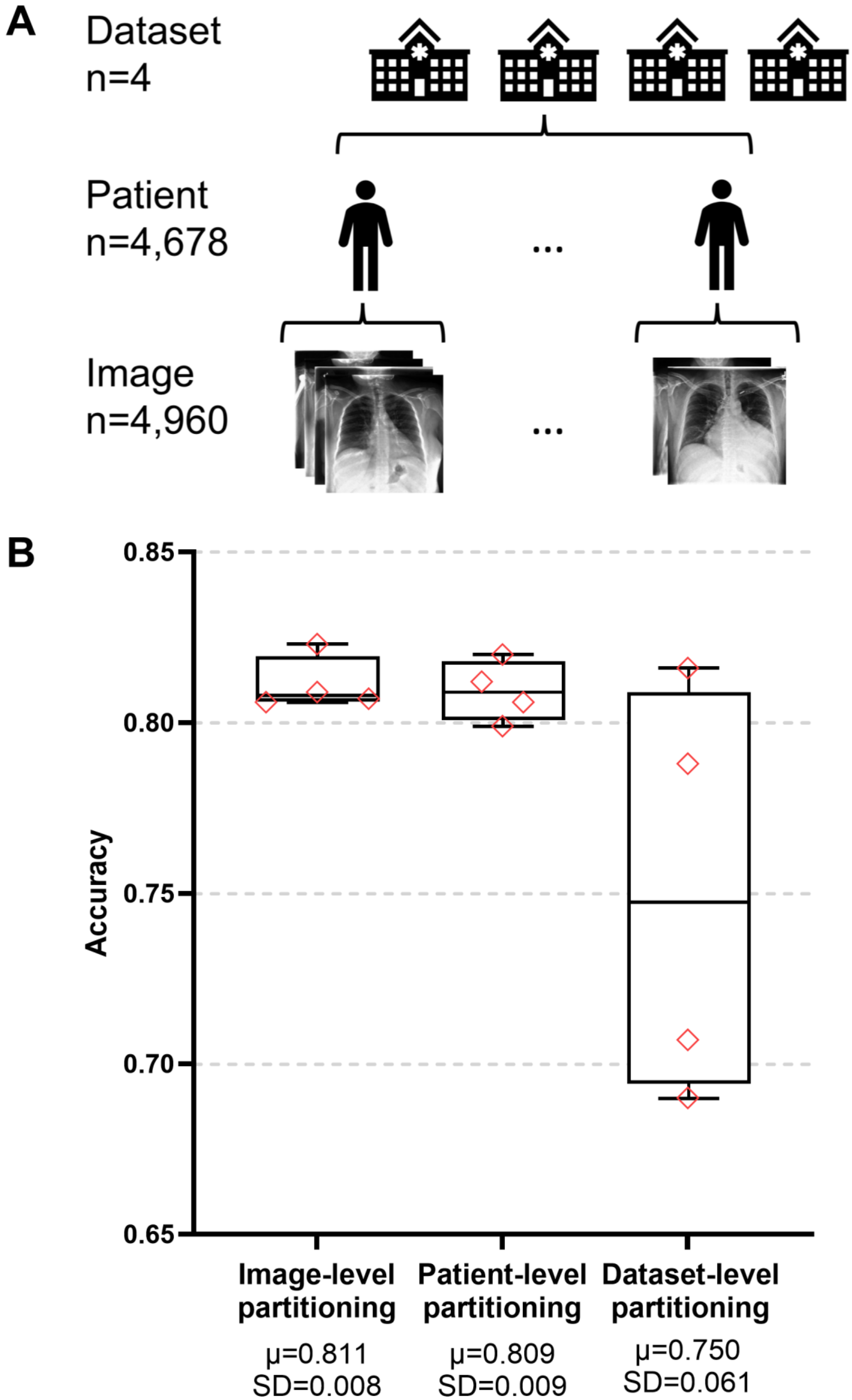

A
Dataset
n=4
Patient
n=4,678
Image
n=4,960
...
...
B
0.85
0.80
0.75
0.70
0.65
Accuracy
Image-level partitioning
Patient-level partitioning
Dataset-level partitioning
μ=0.811
SD=0.008
μ=0.809
SD=0.009
μ=0.750
SD=0.061

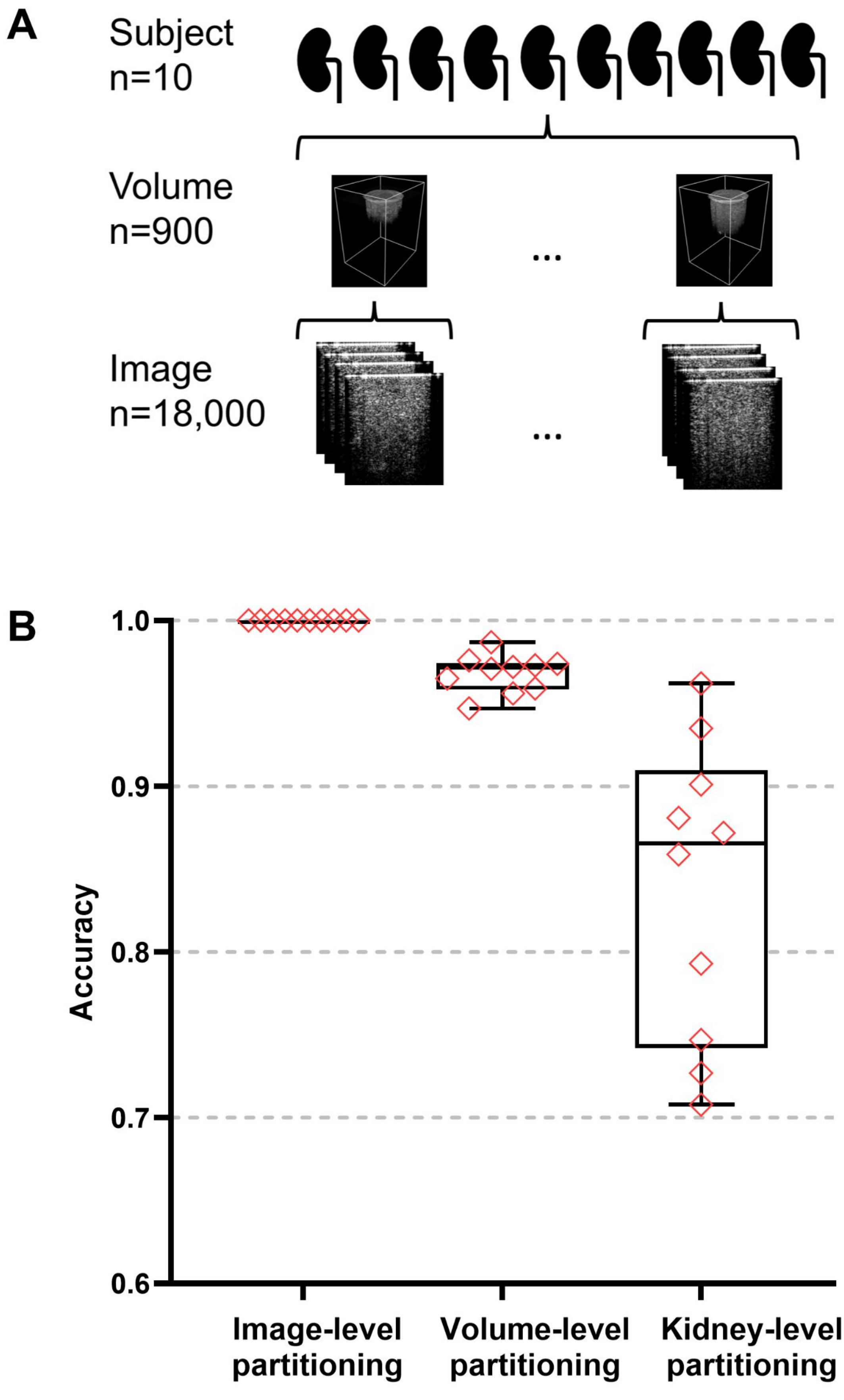
A
Subject
n=10
Volume
n=900
...
Image
n=18,000
...
B
1.0
0.9
0.8
0.7
0.6
Accuracy
Image-level partitioning
Volume-level partitioning
Kidney-level partitioning
μ=1.00
SD=0.00
μ=0.97
SD=0.01
μ=0.84
SD=0.09

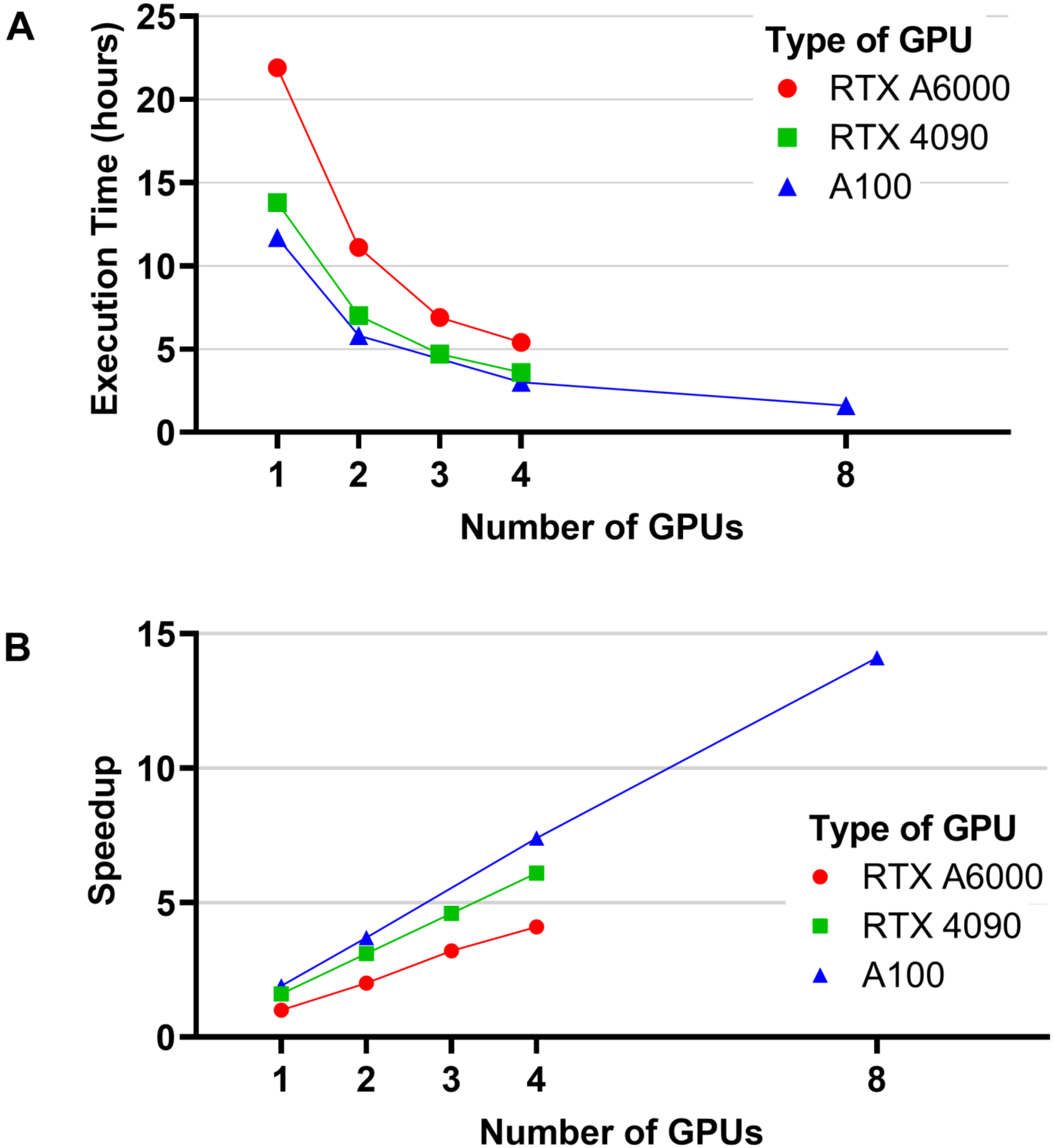

A
Execution Time (hours)
25
20
15
10
5
0
1
2
3
4
8
Number of GPUs
Type of GPU
RTX A6000
RTX 4090
A100
B
Speedup
15
10
5
0
1
2
3
4
8
Number of GPUs
Type of GPU
RTX A6000
RTX 4090
A100

**Algorithm 1:** NACHOS

**Required**

$k$: number of folds

$n$: number of set of hyperparameter configurations from random search

$D$: Dataset

$I = \{0, 1, \ldots k-1\}$

$J = \{0, 1, \ldots n-1\}$

$H = \{h_0, h_1, \ldots h_n\}$

**Defined**

$v_m^j$: Validation performance for hyperparameter configuration $h_j$ on the validation fold $F_m$

$\bar{v}^j$: Average validation performance of hyperparameter configuration $h_j$ across all validation folds

$t_i$: Test performance on the test fold $F_i$

**Result**

average and standard error for performance metric

```
Split D into k folds: F_0, F_1, ... F_{k-1}
/* Cross-testing loop */
for i ∈ I do
    Set F_i as test dataset
    /* AHPO loop */
    for j ∈ J do
        Set h_j as hyperparameter configuration
        /* Cross-validation loop */
        for m ∈ I − {i} do
            Set F_m as validation dataset
            // Distribution of tasks
            Train model on remaining folds (I − {i, m}) with h_j
            v_m^j ← Evaluate model for validation performance on F_m
        v̄^j ← Average values v_m^j, m ∈ I − {i})
    j* ← arg max_j {v̄^j : j ∈ J}
    Train model on folds: I − {i} with h_{j*}
    t_i ← Evaluate model for test performance on F_i
return average and standard error for values t_i, i ∈ I
```

**Algorithm 2:** DACHOS

**Required**

$k$: number of folds

$D$: Dataset

$I = \{0, 1, \dots k - 1\}$

$J = \{0, 1, \dots n - 1\}$

$H = \{h_0, h_1, \dots h_n\}$

**Defined**

$v_m^j$: Validation performance for hyperparameter configuration $h_j$ on the validation fold $F_m$

$\bar{v}^j$: Average validation performance of hyperparameter configuration $h_j$ across all validation folds

**Result**

$M$: Model ready to be deployed

```
Split D into k folds: F_0, F_1, ... F_{k-1}
/* AHPO loop                                         */
for j ∈ J do
    Set h_j as hyperparameter configuration
    /* Cross-validation loop                         */
    for m ∈ I do
        Set F_m as validation dataset
        // Distribution of tasks
        Train model on remaining folds (I − {m}) with h_j
        v_m^j ← Evaluate model for validation performance
          on F_m
    v̄^j ← Average values v_m^j, m ∈ I
j* ← arg max_j{v̄^j : j ∈ J}
M ← Train model on D with h_{j*}
return M
```

Table 1: Randomly generated hyperparameter configurations for AHPO

| Index | Architecture | Batch size | Learning rate | Decay | Momentum | Nesterov |
|---|---|---|---|---|---|---|
| $h_0$ | ResNet50 | 128 | 0.01 | 0.01 | 0.9 | Enabled |
| $h_1$ | InceptionV3 | 16 | 0.001 | 0.001 | 0.9 | Disabled |
| $h_2$ | ResNet50 | 64 | 0.01 | 0.01 | 0.99 | Enabled |
| $h_3$ | Xception | 16 | 0.001 | 0.001 | 0.5 | Disabled |
| $h_4$ | ResNet50 | 64 | 0.01 | 0.01 | 0.5 | Disabled |
| $h_5$ | ResNet50 | 32 | 0.01 | 0.01 | 0.99 | Enabled |
| $h_6$ | ResNet50 | 32 | 0.0001 | 0.0001 | 0.99 | Disabled |
| $h_7$ | ResNet50 | 32 | 0.01 | 0.01 | 0.9 | Enabled |
| $h_8$ | InceptionV3 | 64 | 0.01 | 0.01 | 0.5 | Disabled |

| | **X-ray** | **OCT** |
|---|---|---|
| # Images | 4,960 | 18,000 |
| # Classes | 2 | 3 |
| Partitioning available | Image, Patient, Dataset | Image, Volume, Kidney |
| Image resolution | 224×224 | 185×210 |
| Details | # datasets: 4<br># patients: 4,678<br># images: 4,960 | # kidneys: 10<br># volumes p/kidney p/class: 30<br># images p/volume: 20 |

Table 2: Summary of dataset characteristics for X-ray and OCT.

Table 3: NACHOS accuracy results for chest X-ray repository

| | Hyperparameter configuration | Fold reserved for Test: $F_0$ | $F_1$ | $F_2$ | $F_3$ |
|---|---|---|---|---|---|
| AHPO/Cross-Validation | $h_0$ | $v_1^0$: 0.53 | $v_0^0$: 0.50 | $v_0^0$: 0.50 | $v_0^0$: 0.54 |
| | | $v_2^0$: 0.61 | $v_2^0$: 0.50 | $v_1^0$: 0.49 | $v_1^0$: 0.58 |
| | | $v_3^0$: 0.75 | $v_3^0$: 0.77 | $v_3^0$: 0.50 | $v_2^0$: 0.50 |
| | $h_1$ | $v_1^1$: 0.66 | **$v_0^1$: 0.80** | $v_0^1$: 0.81 | $v_0^1$: 0.80 |
| | | $v_2^1$: 0.68 | **$v_2^1$: 0.74** | $v_1^1$: 0.76 | $v_1^1$: 0.72 |
| | | $v_3^1$: 0.80 | **$v_3^1$: 0.84** | $v_3^1$: 0.71 | $v_2^1$: 0.50 |
| | $h_2$ | **$v_1^2$: 0.69** | $v_0^2$: 0.60 | $v_0^2$: 0.70 | $v_0^2$: 0.67 |
| | | **$v_2^2$: 0.70** | $v_2^2$: 0.67 | $v_1^2$: 0.64 | $v_1^2$: 0.68 |
| | | **$v_3^2$: 0.76** | $v_3^2$: 0.71 | $v_3^2$: 0.66 | $v_2^2$: 0.61 |
| | $h_3$ | $v_1^3$: 0.67 | $v_0^3$: 0.67 | $v_0^3$: 0.71 | $v_0^3$: 0.69 |
| | | $v_2^3$: 0.63 | $v_2^3$: 0.65 | $v_1^3$: 0.63 | $v_1^3$: 0.66 |
| | | $v_3^3$: 0.82 | $v_3^3$: 0.79 | $v_3^3$: 0.76 | $v_2^3$: 0.56 |
| | $h_4$ | $v_1^4$: 0.50 | $v_0^4$: 0.51 | $v_0^4$: 0.50 | $v_0^4$: 0.50 |
| | | $v_2^4$: 0.50 | $v_2^4$: 0.50 | $v_1^4$: 0.50 | $v_1^4$: 0.50 |
| | | $v_3^4$: 0.50 | $v_3^4$: 0.78 | $v_3^4$: 0.67 | $v_2^4$: 0.50 |
| | $h_5$ | $v_1^5$: 0.69 | $v_0^5$: 0.76 | $v_0^5$: 0.70 | **$v_0^5$: 0.73** |
| | | $v_2^6$: 0.69 | $v_2^5$: 0.67 | $v_1^5$: 0.69 | **$v_1^5$: 0.72** |
| | | $v_3^5$: 0.76 | $v_3^5$: 0.76 | $v_3^5$: 0.74 | **$v_2^5$: 0.64** |
| | $h_6$ | $v_1^6$: 0.50 | $v_0^6$: 0.77 | $v_0^6$: 0.75 | $v_0^6$: 0.73 |
| | | $v_2^6$: 0.51 | $v_2^6$: 0.66 | $v_1^6$: 0.50 | $v_1^6$: 0.50 |
| | | $v_3^6$: 0.77 | $v_3^6$: 0.76 | $v_3^6$: 0.78 | $v_2^6$: 0.50 |
| | $h_7$ | $v_1^7$: 0.70 | $v_0^7$: 0.77 | $v_0^7$: 0.74 | $v_0^7$: 0.51 |
| | | $v_2^7$: 0.49 | $v_2^7$: 0.69 | $v_1^7$: 0.67 | $v_1^7$: 0.69 |
| | | $v_3^7$: 0.78 | $v_3^7$: 0.80 | $v_3^7$: 0.74 | $v_2^7$: 0.56 |
| | $h_8$ | $v_1^8$: 0.50 | $v_0^8$: 0.80 | **$v_0^8$: 0.81** | $v_0^8$: 0.81 |
| | | $v_2^8$: 0.61 | $v_2^8$: 0.71 | **$v_1^8$: 0.74** | $v_1^8$: 0.73 |
| | | $v_3^8$: 0.78 | $v_3^8$: 0.80 | **$v_3^8$: 0.77** | $v_2^8$: 0.50 |
| | best: $h_{j^*}$ | $h_2$ | $h_1$ | $h_8$ | $h_5$ |
| | Average | $\bar{v}^2$: 0.72 | $\bar{v}^1$: 0.79 | $\bar{v}^8$: 0.77 | $\bar{v}^5$: 0.69 |
| Cross-Testing | Accuracy | $t_0$: 0.79 | $t_1$: 0.71 | $t_2$: 0.69 | $t_3$: 0.82 |
| | Average and standard error 0.75±0.03 | | | | |

Note: $F_i$: represents the test fold $i$. $h_j$: represents the hyperparameter configuration $j$. In the AHPO/CV loop, inside each cell, the validation accuracy $v_m^j$ for hyperparameter configuration $j$ and validation fold $m$ is located. For each test fold, the best hyperparameter configuration is selected by comparing the average validation accuracy and the whole cell is highlighted in blue. For instance, for $F_0$, the hyperparameter configuration $h_2$ has the highest average accuracy. $\bar{v}^2$=(0.69+0.70+0.76)/3=0.72. In the cross-testing loop, the selected hyperparameter configuration is used to calculate the test accuracy $t_i$ for test fold $i$. For test fold $F_0$, test accuracy is $t_0$=0.79

Table 4: NACHOS accuracy results for kidney OCT dataset

| | Hyperparameter configuration | Fold reserved for Test $F_0$ | $F_1$ | $F_2$ | $F_3$ | $F_4$ | $F_5$ | $F_6$ | $F_7$ | $F_8$ | $F_9$ |
|---|---|---|---|---|---|---|---|---|---|---|---|
| AHPO/ Cross-Validation | $h_0$ | 0.91±0.02 | 0.87±0.04 | 0.83±0.06 | 0.85±0.04 | 0.86±0.03 | 0.86±0.03 | **0.87±0.03** | 0.85±0.03 | 0.85±0.03 | 0.87±0.03 |
| | $h_1$ | 0.92±0.01 | 0.86±0.04 | 0.87±0.03 | 0.83±0.06 | 0.85±0.06 | 0.88±0.04 | 0.83±0.04 | 0.87±0.03 | 0.85±0.03 | 0.87±0.04 |
| | $h_2$ | 0.89±0.02 | 0.87±0.03 | **0.91±0.02** | 0.88±0.02 | 0.85±0.04 | **0.89±0.02** | 0.82±0.04 | **0.90±0.02** | 0.88±0.02 | **0.89±0.03** |
| | $h_3$ | 0.89±0.03 | 0.84±0.05 | 0.85±0.05 | 0.82±0.06 | 0.83±0.06 | 0.80±0.06 | 0.81±0.06 | 0.83±0.04 | 0.83±0.06 | 0.84±0.05 |
| | $h_4$ | 0.86±0.03 | 0.88±0.04 | 0.85±0.04 | 0.81±0.04 | 0.83±0.04 | 0.85±0.03 | 0.84±0.03 | 0.83±0.03 | 0.80±0.03 | 0.83±0.04 |
| | $h_5$ | **0.93±0.01** | **0.89±0.03** | 0.89±0.02 | **0.90±0.02** | **0.88±0.03** | 0.89±0.02 | 0.82±0.03 | 0.90±0.02 | 0.87±0.02 | 0.87±0.03 |
| | $h_6$ | 0.88±0.02 | 0.86±0.04 | 0.89±0.03 | 0.85±0.06 | 0.82±0.05 | 0.88±0.03 | 0.85±0.04 | 0.83±0.03 | 0.85±0.03 | 0.87±0.04 |
| | $h_7$ | 0.89±0.01 | 0.87±0.04 | 0.87±0.03 | 0.86±0.03 | 0.87±0.03 | 0.88±0.03 | 0.87±0.03 | 0.83±0.02 | 0.80±0.03 | 0.88±0.03 |
| | $h_8$ | 0.91±0.02 | 0.89±0.04 | 0.89±0.03 | 0.86±0.04 | 0.86±0.04 | 0.88±0.03 | 0.85±0.05 | 0.88±0.04 | **0.89±0.03** | 0.88±0.04 |
| | best: $h_{j^*}$ | $h_5$ | $h_5$ | $h_2$ | $h_5$ | $h_5$ | $h_2$ | $h_0$ | $h_2$ | $h_8$ | $h_2$ |
| Cross-Testing | Accuracy | $t_0$: 0.73 | $t_1$: 0.71 | $t_2$: 0.79 | $t_3$: 0.75 | $t_4$: 0.88 | $t_5$: 0.87 | $t_6$: 0.94 | $t_7$: 0.90 | $t_8$: 0.96 | $t_9$: 0.86 |
| | | Average and standard error 0.84±0.03 | | | | | | | | | |

Note: $F_i$: represents the test fold $i$. $h_j$: represents the hyperparameter configuration $j$. In the AHPO/CV loop, inside each cell, the average and standard error of the validation accuracies $v_m^j$ for hyperparameter configuration $j$ and validation fold $m$ are located. The cells highlighted in blue have the highest average accuracy for a test fold. For instance, for $F_0$, the hyperparameter configuration $h_5$ has the highest average accuracy. $\bar{v}^5$=0.93. In the cross-testing loop, the selected hyperparameter configuration is used to calculate the test accuracy $t_i$.

Table 5: Average confusion matrix for chest X-ray classification Cross-Testing loop NACHOS

| | | Predicted | |
|---|---|---|---|
| | | No-Finding | Cardiomegaly |
| Truth | No-Finding | 426±82 | 194±82 |
| | Cardiomegaly | 115±66 | 505±66 |

Note: average and standard deviation. Ground Truth: 612 No-Finding and 612 Cardiomegaly

Table 6: Average confusion matrix for chest OCT classification Cross-Testing loop NACHOS

| | | Predicted | | |
|---|---|---|---|---|
| | | Cortex | Medulla | Pelvis |
| Truth | Cortex | 454±136 | 40±58 | 106±147 |
| | Medulla | 80±136 | 512±137 | 8±19 |
| | Pelvis | 55±72 | 2±6 | 543±76 |

Note: average and standard deviation. Ground Truth: 600 Cortex, 600 Medulla, and 600 Pelvis

Table 7: DACHOS accuracy results for chest X-ray repository

| Hyperparameter configuration | Fold reserved for Validation | | | | |
|---|---|---|---|---|---|
| | $F_0$ | $F_1$ | $F_2$ | $F_3$ | Average |
| $h_0$ | $v_0^0$: 0.71 | $v_1^0$: 0.71 | $v_2^0$: 0.69 | $v_3^0$: 0.70 | $\bar{v}^0$: 0.71 |
| $h_1$ | $v_0^1$: 0.79 | $v_1^1$: 0.63 | $v_2^1$: 0.71 | $v_3^1$: 0.71 | $\bar{v}^1$: 0.71 |
| $h_2$ | $v_0^2$: 0.78 | $v_1^2$: 0.70 | $v_2^2$: 0.69 | $v_3^2$: 0.65 | $\bar{v}^2$: 0.71 |
| $h_3$ | $v_0^3$: 0.75 | $v_1^3$: 0.69 | $v_2^3$: 0.66 | $v_3^3$: 0.80 | $\bar{v}^3$: 0.72 |
| $h_4$ | $v_0^4$: 0.50 | $v_1^4$: 0.50 | $v_2^4$: 0.59 | $v_3^4$: 0.78 | $\bar{v}^4$: 0.59 |
| $h_5$ | $v_0^5$: 0.75 | $v_1^5$: 0.74 | $v_2^5$: 0.71 | $v_3^5$: 0.80 | **$\bar{v}^5$: 0.75** |
| $h_6$ | $v_0^6$: 0.73 | $v_1^6$: 0.71 | $v_2^6$: 0.73 | $v_3^6$: 0.77 | $\bar{v}^6$: 0.73 |
| $h_7$ | $v_0^7$: 0.77 | $v_1^7$: 0.72 | $v_2^7$: 0.63 | $v_3^6$: 0.75 | $\bar{v}^7$: 0.72 |
| $h_8$ | $v_0^8$: 0.81 | $v_1^8$: 0.74 | $v_2^8$: 0.63 | $v_3^8$: 0.81 | $\bar{v}^8$: 0.75 |
| best: $h_{j^*}$ | $h_5$ | | | | |

Note: $F_m$: represents the validation fold $m$. $h_j$: represents the hyperparameter configuration $j$. Inside each cell, the validation performance $v_m^j$ for hyperparameter configuration $j$ and validation fold $m$ is located. In order to select the configuration hyperparameter, the highest average accuracy $\bar{v}^j$ is selected. Among them, $h_5$ has the highest average validation accuracy $\bar{v}^5$=0.75, highlighted in blue. Despite apparent equality due to rounding, $h_5$'s average validation accuracy is higher than $h_8$.

Table 8: DACHOS accuracy results for kidney OCT dataset

| Hyperparameter configuration | Fold reserved for Validation | | | | | | | | | | |
|---|---|---|---|---|---|---|---|---|---|---|---|
| | $F_0$ | $F_1$ | $F_2$ | $F_3$ | $F_4$ | $F_5$ | $F_6$ | $F_7$ | $F_8$ | $F_9$ | Average |
| $h_0$ | 0.63 | 0.88 | 0.85 | 0.76 | 0.91 | 0.97 | 0.92 | 0.87 | 0.94 | 0.88 | $\bar{v}^0$: 0.86 |
| $h_1$ | 0.51 | 0.91 | 0.79 | 0.93 | 0.90 | 0.91 | 0.98 | 0.84 | 0.97 | 0.90 | $\bar{v}^1$: 0.86 |
| $h_2$ | 0.77 | 0.88 | 0.81 | 0.91 | 0.96 | 0.93 | 0.92 | 0.91 | 0.97 | 0.92 | **$\bar{v}^2$: 0.90** |
| $h_3$ | 0.37 | 0.90 | 0.82 | 0.83 | 0.87 | 0.91 | 0.91 | 0.92 | 0.98 | 0.87 | $\bar{v}^3$: 0.84 |
| $h_4$ | 0.68 | 0.89 | 0.86 | 0.84 | 0.78 | 0.99 | 0.88 | 0.86 | 0.93 | 0.91 | $\bar{v}^4$: 0.86 |
| $h_5$ | 0.77 | 0.88 | 0.89 | 0.79 | 0.94 | 0.97 | 0.92 | 0.91 | 0.96 | 0.93 | $\bar{v}^5$: 0.90 |
| $h_6$ | 0.66 | 0.88 | 0.81 | 0.91 | 0.84 | 0.95 | 0.91 | 0.90 | 0.97 | 0.92 | $\bar{v}^6$: 0.88 |
| $h_7$ | 0.69 | 0.87 | 0.84 | 0.93 | 0.85 | 0.79 | 0.95 | 0.86 | 0.97 | 0.91 | $\bar{v}^7$: 0.87 |
| $h_8$ | 0.64 | 0.89 | 0.69 | 0.94 | 0.92 | 0.92 | 0.97 | 0.90 | 0.97 | 0.90 | $\bar{v}^8$: 0.88 |
| best: $h_{j^*}$ $h_2$ | | | | | | | | | | | |

Note: $F_m$: represents the validation fold $m$. $h_j$: represents the hyperparameter configuration $j$. Inside each cell, the validation accuracy $v_m^j$ for hyperparameter configuration $j$ and validation fold $m$ is located. In order to select the configuration hyperparameter, the highest average accuracy $\bar{v}^j$ is selected. Among them, $h_2$ has the highest average validation accuracy $\bar{v}^2$=0.90, highlighted in blue. Despite apparent equality due to rounding, $h_2$'s average validation accuracy is higher than $h_5$.

## S1. Implementation Details and Hardware Configuration

NACHOS and DACHOS were implemented using Python 3.8 and TensorFlow 2.6.2. They use `dill` for saving and loading configuration checkpoints, `mpi4py` for parallelization, `fasteners` for process locking and unlocking, `NumPy` for working with arrays, `scikit-learn` for computing performance metrics, `SciPy` for statistical analysis of the results, and `termcolor` for color-coded standard output messages. NACHOS and DACHOS are capable of generating learning curves, confusion matrices, and Receiver Operating Characteristic (ROC) curves for result visualization. They can also generate class activation maps for instance-wide prediction interpretation or feature importance, using GradCAM, which requires `Matplotlib` and `seaborn`.

NACHOS and DACHOS were designed to operate on both a supercomputer with GPU nodes and a Beowulf cluster of GPU workstations connected via a local Ethernet network. The algorithms were tested on the Schooner supercomputer, utilizing GPU nodes equipped with NVIDIA A100 GPUs. Jobs on the supercomputer were managed through the SLURM system, with the GPU count per node specified. Additional experiments were performed on a Beowulf cluster comprising GPU workstations with NVIDIA RTX A6000 and NVIDIA RTX 4090 GPUs, running Ubuntu 20.04.6 LTS. In the Beowulf cluster, data were distributed across all workstations, and jobs were executed using a configuration file that specified the GPU count and the IP address of each workstation. The repository for this paper can be fount at `https://github.com/thepanlab/NACHOS`.

## S2. Tables

Table S1: Performance comparison across different GPU types and configurations.

| GPU type | Memory bandwidth | # tensor cores | # GPUs | Time (hrs) |
|---|---|---|---|---|
| RTXA6000 | 112.5 GB/s | 336 | 1 | 21.9 |
| RTXA6000 | 112.5 GB/s | 336 | 2 | 11.1 |
| RTXA6000 | 112.5 GB/s | 336 | 3 | 6.9 |
| RTXA6000 | 112.5 GB/s | 336 | 4 | 5.4 |
| RTX4090 | 1000.0 GB/s | 512 | 1 | 13.8 |
| RTX4090 | 1000.0 GB/s | 512 | 2 | 7.0 |
| RTX4090 | 1000.0 GB/s | 512 | 3 | 4.7 |
| RTX4090 | 1000.0 GB/s | 512 | 4 | 3.6 |
| RTXA6000 + RTX4090 | - | - | 8 | 2.2 |
| A100 | 2000.0 GB/s | 432 | 1 | 11.7 |
| A100 | 2000.0 GB/s | 432 | 2 | 5.8 |
| A100 | 2000.0 GB/s | 432 | 4 | 3.0 |
| A100 | 2000.0 GB/s | 432 | 8 | 1.6 |